\documentclass[a4paper,fleqn]{cas-dc}

\usepackage[numbers]{natbib}

\usepackage[utf8]{inputenc}

\usepackage[english]{babel}
\usepackage[T1]{fontenc}

\usepackage{amsmath}
\usepackage{amsfonts}
\usepackage{graphicx}
\usepackage{stfloats}
\usepackage[colorinlistoftodos]{todonotes}

\usepackage{indentfirst}
\usepackage{algorithm}          
\usepackage{algorithmic}

\usepackage{amsthm}
\newtheorem{definition}{Definition}
\theoremstyle{definition}
\newcommand{\blob}{\rule[.2ex]{1ex}{1ex}}
\newcommand{\squishlist}{ 
   \begin{list}{$\bullet$}
    { \setlength{\itemsep}{0pt}      \setlength{\parsep}{3pt} 
      \setlength{\topsep}{3pt}       \setlength{\partopsep}{0pt}
      \setlength{\leftmargin}{1.5em} \setlength{\labelwidth}{1em}
      \setlength{\labelsep}{0.5em} } }

\newcommand{\squishend}{
    \end{list}  } 

\usepackage{float}

\def\tsc#1{\csdef{#1}{\textsc{\lowercase{#1}}\xspace}}
\tsc{WGM}
\tsc{QE}
\tsc{EP}
\tsc{PMS}
\tsc{BEC}
\tsc{DE}
\begin{document} 
\let\WriteBookmarks\relax
\def\floatpagepagefraction{1}
\def\textpagefraction{.001}
\shorttitle{\textit{I-MAD: Interpretable Malware Detector}}
\shortauthors{M. Li et~al.}
\title [mode = title]{\textit{I-MAD}: Interpretable Malware Detector Using Galaxy Transformer}      

\author[1]{Miles Q. Li}
\ead{miles.qi.li@mail.mcgill.ca}

\credit{Conceptualization, Methodology, Software, Validation, Data Curation, Investigation, Writing - Original Draft}

\author[2]{Benjamin C. M. Fung}[orcid=0000-0001-8423-2906]\cormark[1]
\ead{ben.fung@mcgill.ca}

\credit{Conceptualization, Validation, Supervision, Writing - Review \& Editing, Funding acquisition}

\author[3]{Philippe Charland}
\ead{philippe.charland@drdc-rddc.gc.ca}

\credit{Project administration, Writing - Review \& Editing}

\author[4]{Steven H. H. Ding}
\ead{ding@cs.queensu.ca}

\credit{Data Curation, Writing - Review \& Editing}

\address[1]{School of Computer Science, McGill University, Montreal, Canada}

\address[2]{School of Information Studies, McGill University, Montreal, Canada}

\address[3]{Mission Critical Cyber Security Section, Defence R\&D Canada, Quebec, Canada}

\address[4]{School of Computing, Queen's University, Kingston, Canada}

\cortext[cor1]{Corresponding author.}

\begin{abstract}
Malware currently presents a number of serious threats to computer users. Signature-based malware detection methods are limited in detecting new malware samples that are significantly different from known ones. Therefore, machine learning-based methods have been proposed, but there are two challenges these methods face. The first is to model the full semantics behind the assembly code of malware. The second challenge is to provide interpretable results while keeping excellent detection performance. In this paper, we propose an \textit{Interpretable MAlware Detector} (\textit{I-MAD}) that outperforms state-of-the-art static malware detection models regarding accuracy with excellent interpretability. To improve the detection performance, \textit{I-MAD} incorporates a novel network component called the \textit{Galaxy Transformer network} that can understand assembly code at the basic block, function, and executable levels. It also incorporates our proposed interpretable feed-forward neural network to provide interpretations for its detection results by quantifying the impact of each feature with respect to the prediction. Experiment results show that our model significantly outperforms existing state-of-the-art static malware detection models and presents meaningful interpretations.
\end{abstract}


\maketitle

\section{Introduction}

Malware is software written in order to steal credentials of computer users, damage computer systems, encrypt documents for ransom, and other nefarious goals. Recognizing malware samples downloaded by legitimate users in a timely manner is of crucial importance for users' protection. Signature-based malware detection methods are widely used in antivirus products, but they are limited in recognizing significant variants of existing malware and new malware~\cite{ye2017survey,demontis2017yes}. There is thus a pressing need to create an intelligent malware detection system that has better generability to capture new malware or nontrivial variants of known malware.

Machine learning-based malware analysis methods~\cite{ye2017survey,moskovitch2008unknown,dai2009efficient,baldangombo2013static,saxe2015deep,devlin2018bert} can automatically learn common patterns of malware from the feature space that have better generalization ability than manually crafted signatures. However, there are two major challenges for machine learning-based malware detection models. 

Interpretability is one of the dominant features for classification models in some domains, such as healthcare and cybersecurity. In cybersecurity, the interpretations can help malware analysts justify the classification results and create a knowledge base of malware samples. Hidden Markov model (HMM)~\cite{vemparala2016malware,wong2006hunting} and attention-based recurrent neural network (RNN)~\cite{choi2016retain} have been proposed to provide analyzable or interpretable classification results on sequential data. Linear models such as logistic/softmax regression and Naive Bayes produce interpretable results on vectorial data but usually yield inferior classification performance than non-linear models such as multi-layer feed-forward neural networks~\cite{cerna2019interpretable}. However, the hidden layers between the input and the logistic/softmax layer make multi-layer feed-forward neural networks lose the interpretability of logistic/softmax regression to directly attribute the impact of each feature. It's still a challenge to keep interpretability as well as  classification performance for feed-forward neural networks.

As the workload of malware exists mainly in its assembly code, modelling the assembly code could provide important information for malware detection. However, it is challenging to model the whole assembly code of executables because they are very long sequences. An executable of 1 MB could have hundreds of thousands of instructions. No effective training approaches have been proposed to train such long sequences, and the memory consumption cannot be handled with standard hardware for such long sequences. 

Deep learning models have achieved significant breakthroughs in understanding natural language when properly trained on large corpora~\cite{radford2018improving,devlin2018bert,radford2019language}.  \textit{Transformer}~\cite{vaswani2017attention} based models especially achieve state-of-the-art results in natural language understanding and generation~\cite{devlin2018bert,radford2018improving,Dong2019UnifiedLM,radford2019language,raffel2019exploring,brown2020language}. However, their successful applications are mainly on short text, i.e., sentence-level tasks such as paraphrase detection and sentiment analysis~\cite{radford2018improving,devlin2018bert}, or on  short-document texts such as reading comprehension and automatic summarization of news articles~\cite{Dong2019UnifiedLM}. For example, the state-of-the-art sequence model \textit{GPT-3}~\cite{brown2020language} can process sequences of a maximum length of 2,048 tokens. That makes the transference of the success of existing methods to understanding assembly code a challenge. Apart from the fact that assembly code is too long, the differences between natural language and assembly code in the structure composition and basic units stand as another problem to solve.

\begin{figure*}
\centering
\includegraphics[width=16cm]{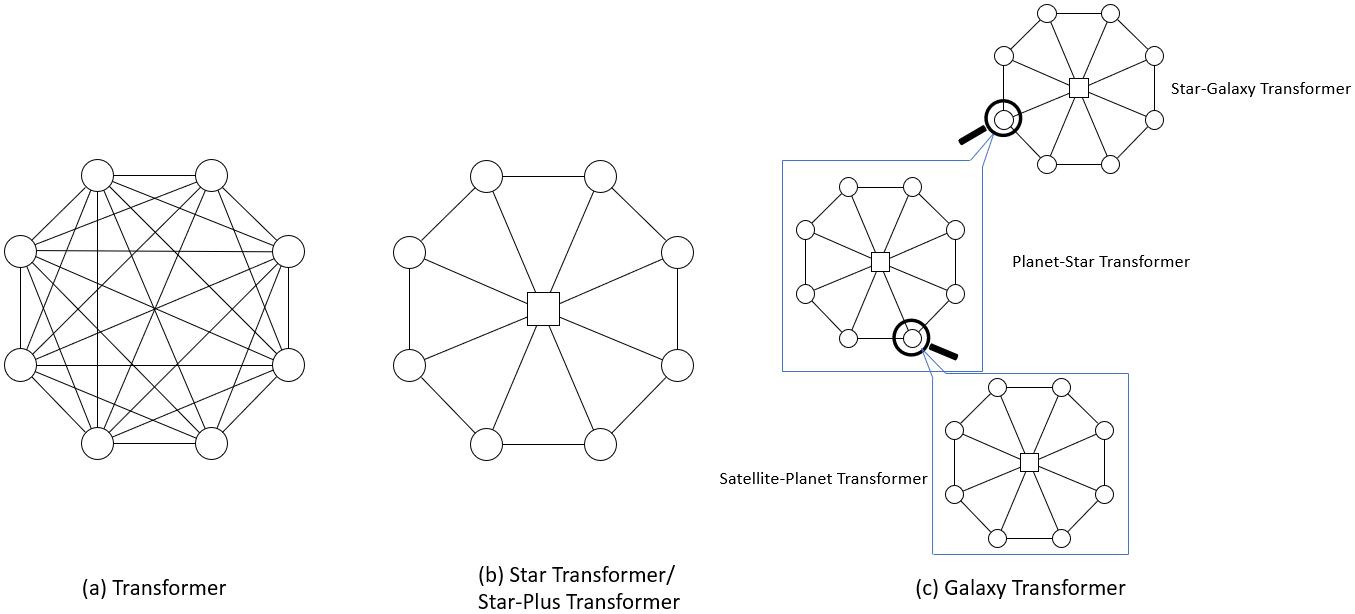}\vspace{-2mm}
 \caption{The comparison of topology of the \textit{Transformer}, \textit{Star/Star-Plus Transformer}, and \textit{Galaxy Transformer}.}\vspace{-5mm}
 \label{fig:topologies}
\end{figure*}

Despite the fact that the assembly code of an executable is usually very long, it has an innate hierarchical structure: instructions form basic blocks, basic blocks form assembly functions, and assembly functions form the ensemble of assembly code (i.e., the full logic) of an executable. The lengths of basic blocks, assembly functions, and the ensemble of the assembly code of an executable in terms of their direct sub-units are usually within thousands. Based on this characteristic, we propose the \textit{Galaxy Transformer} network. It contains three components, namely the \textit{Satellite-Planet Transformer}, the \textit{Planet-Star Transformer}, and the \textit{Star-Galaxy Transformer}. They are three customized \textit{Star-Plus Transformer} networks organized in a hierarchy in order to understand the semantic meaning of the assembly code of an executable at different levels: basic block, assembly function, and executable. The \textit{Star-Plus Transformer} is our improved version of the \textit{Star Transformer}~\cite{guo2019star}, which was proposed for natural language understanding as a variant of the \textit{Transformer}~\cite{vaswani2017attention}. The time complexity and space complexity of \textit{Transformer} is $O(n^2)$, where $n$ is the length of the token. The \textit{Star Transformer} replaces the fully connected structure of the \textit{Transformer} with a star-shaped topology to reduce the complexities to $O(n)$, and it achieves better results on modestly sized datasets. A comparison of the topology between the \textit{Transformer}, the \textit{Star/Star-Plus Transformer}, and the \textit{Galaxy Transformer} is shown in Figure~\ref{fig:topologies}. Our proposed universe-like topology of the \textit{Galaxy Transformer} makes it possible to train very long sequences. 

To provide interpretations for the classification results, we propose a novel \textit{interpretable feed-forward neural network} (\textit{IFFNN}) as the other key component of our full model, the \textit{Interpretable MAlware Detector} (\textit{I-MAD}). It has the modelling power of a multi-layer neural network and the interpretability of a logistic regression model. An example of the prediction and its interpretation is given in Table~\ref{tab:example}. It shows the detection result of a target file, the confidence in the result, the primary contributing features that lead to the prediction, and the most related assembly functions.

The contributions of this paper are summarized below:
\begin{enumerate}
\item We propose the \textit{Galaxy Transformer} as an early attempt in the literature to model the full sequences of  assembly code for malware detection. 
\item We propose two pre-training tasks to train the \textit{Satellite-Planet Transformer} and \textit{Planet-Star Transformer}, which are both components of the \textit{Galaxy Transformer}, to understand the semantic meaning of assembly code at the basic block and assembly function levels.
\item We improve the way to use printable string features and PE import features from previous works with our insights on malware. 
\item We propose a novel IFFNN as the classification module of \textit{I-MAD}. It has the same interpretability as logistic regression and the modelling power of multi-layer feed-forward neural networks. It allows \textit{I-MAD} to quantify the impact of each feature for the classification results. It is also a general classification module that can be applied to other classification tasks.
\end{enumerate}

The rest of the paper is organized as follows. Section~\ref{sec_related} discusses related work. Section~\ref{sec_probdef} defines the research problem. Section~\ref{sec_arc} provides the details of our proposed method. Section~\ref{sec_exp} presents the experiment results and analyses. Section~\ref{sec_lim} discusses the limitations and our future work. Section~\ref{sec_con} concludes the paper.

\section{Related Work}
\label{sec_related}
\subsection{Malware Detection}
Malware detection methods fall into three categories: static, dynamic, and hybrid~\cite{damodaran2017comparison}. We summarize the common static and dynamic features in Table~\ref{tab:stadyfeatures}.

Static methods examine the static content of an executable, while dynamic methods run an executable and analyze its behaviors. Features used in static methods include binary sequences~\cite{schultz2001data,kolter2004learning,anderson2012improving,saxe2015deep,raff2017malware,gibert2020hydra}, assembly code sequences~\cite{moskovitch2008unknown,dai2009efficient,anderson2011graph,anderson2012improving,santos2013opem,gibert2020hydra}, numerical PE header features~\cite{anderson2012improving,baldangombo2013static,saxe2015deep}, PE imports/API calls~\cite{schultz2001data,baldangombo2013static,saxe2015deep,onwuzurike2019mamadroid,gibert2020hydra}, printable strings~\cite{schultz2001data,islam2013classification,saxe2015deep}, and malware images~\cite{mourtaji2019intelligent,verma2020multiclass,vasan2020image}. The most common way to use binary sequences or assembly code sequences is to cut them into n-gram pieces to form features~\cite{schultz2001data,kolter2004learning,moskovitch2008unknown,dai2009efficient,anderson2011graph,anderson2012improving,santos2013opem,saxe2015deep}. Some studies find that byte n-grams are effective features~\cite{basole2020multifamily}, while others suggest that byte n-grams are weak or deeply flawed~\cite{raff2018investigation} and assembly code is more effective~\cite{moskovitch2008unknown,santos2013opem}. As control flow graphs could be more robust than assembly code against some obfuscation techniques, there are also instance-based detection methods that identify malware by checking whether an executable contains assembly functions or control flow graphs of known malware~\cite{cesare2010classification,anderson2012improving,cesare2013control,chen2015detecting}. Because they are instance-based methods, they suffer from efficiency issues when the known malware database is large.

Dynamic methods run a target executable in an isolated environment, e.g., a virtual machine or an emulator, and extract features such as the memory image~\cite{kruegel2005polymorphic,dahl2013large,huang2016mtnet}, the executed instructions~\cite{royal2006polyunpack,dai2009efficient,anderson2011graph,anderson2012improving}, and the invoked system calls or behaviors derived from them~\cite{bayer2006dynamic,fredrikson2010synthesizing,anderson2012improving,dahl2013large,islam2013classification,santos2013opem,huang2016mtnet,amer2020dynamic}. 

\begin{table}[t]
\caption{Sample result of our malware detection and its interpretation, which includes the 5 factors that contribute most to the prediction and the most related assembly functions.}
\label{tab:example}
\begin{center}
\begin{tabular}{|c|c|c|}
\hline
\multicolumn{3}{|c|}{File: 05c199.exe}\\\hline
\multicolumn{3}{|c|}{Prediction: malicious}\\\hline
\multicolumn{3}{|c|}{Confidence: 100\%}\\\hline\hline
\multicolumn{3}{|c|}{Primary factors leading to the prediction of malicious}\\\hline
Feature description&Feature value&Impact\\\hline
Assembly code&N/A&14.56\\\hline
Number of PE imports&8&5.12\\\hline
Major operating system version&1&1.49\\\hline
Frequency of the string "Sleep"&1&0.82\\\hline
Frequency of the string ".data"&1&0.59\\\hline
\hline
\multicolumn{3}{|c|}{Most influential assembly functions}\\\hline
\multicolumn{3}{|c|}{sub\_401010}\\\hline
\multicolumn{3}{|c|}{sub\_4010AE}\\\hline
\end{tabular}\vspace{-5mm}
\end{center}
\end{table}
Both static and dynamic methods have their advantages and disadvantages. Compared with static methods, dynamic methods provide more abundant and direct information. Even though both static and dynamic methods extract system calls as features, the parameters passed to those invoked system calls can always be seen with dynamic methods, which is not the case with static methods~\cite{bayer2006dynamic,dahl2013large,islam2013classification,santos2013opem}. Moreover, when a malicious executable is packed or polymorphic, the payload probably cannot be seen by static methods. Yet, to perform its malicious actions it must reveal the payload during execution~\cite{bayer2006dynamic}. This gives another advantage to dynamic over static methods. Therefore, dynamic methods can often achieve better results in the most challenging cases~\cite{vemparala2016malware}. However, it does not mean that static methods cannot capture malware with those mechanisms, because their use is suspicious and can be detected. Previous works on static malware detection show that when analyzing an unknown executable from multiple feature scopes, it is hard for the malware to evade detection~\cite{anderson2012improving,islam2013classification}. On the other hand, one serious shortcoming of dynamic methods is that when malware finds that its execution is being monitored, it may not perform its malicious action to evade detection. Thus, dynamic methods may fail to detect it~\cite{bayer2006dynamic,ye2017survey}. In addition, dynamically analyzing an executable is very time consuming. 

Hybrid methods extract both static and dynamic features and integrate them into one malware detection model~\cite{anderson2012improving,islam2013classification,santos2013opem,damodaran2017comparison}. These two kinds of features are expected to provide complementary information to the model so that it has a more comprehensive view of a sample.

\begin{table}[h]
\caption{Common static and dynamic features for malware detection.}
\label{tab:stadyfeatures}
\begin{center}
\begin{tabular}{|c|c|}
\hline
Static&Dynamic\\
\hline
binary sequences&memory image\\
assembly code&executed instructions\\
PE header numerical fields&invoked system calls\\
PE imports/API calls&behaviors\\
printable strings&\\
malware images&\\
control flow graph&\\
\hline
\end{tabular}\vspace{-5mm}
\end{center}
\end{table}

\subsection{Transformers}
As programming languages and natural languages share some similar characteristics, the experience in modeling the latter can be customized to model the former. Before Vaswani et al.~\cite{vaswani2017attention} proposed the deep learning model known as the \textit{Transformer}, most state-of-the-art neural machine translation models belonged to the class of \textit{attention-based recurrent neural network} (\textit{RNN}) models. In these models, an \textit{RNN} is used to encode the source text, and another \textit{RNN} with attention mechanism is used to generate the translation word by word~\cite{bahdanau2014neural,luong2015effective}. The attention mechanism is used to determine the importance of the words in the source text for generating each translated word. One disadvantage of this type of model is that the recurrence nature precludes parallelism. Another disadvantage is that the attention mechanism assigns only one importance weight to a word in the source text so it can focus on just one aspect of the words. 

The \textit{Transformer} addresses both problems and achieves new state-of-the-art performance on machine translation by abandoning the \textit{RNN} and relying only on an improved attention mechanism~\cite{vaswani2017attention}. The attention mechanism in the \textit{Transformer} is referred to as \textit{multi-head attention}, which allows multiple attention weights to be assigned to each item. Each weight corresponds to one aspect of an item, thus their attention mechanism is more powerful than the previously proposed attention mechanism in its modeling ability~\cite{vaswani2017attention}. As there is no \textit{RNN} in it, the recurrence nature of the encoder does not exist anymore, which tremendously increases the parallelism and computing efficiency. Since 2017, researchers have seen the potential of the \textit{Transformer} and proposed their own ways to pre-train the \textit{Transformer} on unlabeled corpora that are abundant and then fine-tune it for downstream NLP tasks. They constantly achieve significantly better results than previous methods on many NLP tasks~\cite{radford2018improving,devlin2018bert,Dong2019UnifiedLM,radford2019language,yang2019xlnet,raffel2019exploring,brown2020language}. 
The problem of the \textit{Transformer} is that it computes the attention weights between any two items of a sequence, which leads to $O(n^2)$ time and space complexity. Therefore, the \textit{Star Transformer}~\cite{guo2019star} is proposed to reduce the complexities to $O(n)$ by adding an additional node to collect global information and connect only adjacent items of the sequence. When the length of a sequence is very long, such a sequential model is still hard to train. For this reason, we propose the \textit{Galaxy Transformer} with a hierarchical topology, so that it has $O(n)$ complexities and can be trained at different levels.

\subsection{Interpretable Networks}
In most cases, deep learning models are proposed to achieve the best performance for certain research problems without considering their interpretability. However, interpretability is very important in some fields. In healthcare, the rationale for decisions or predictions made by deep learning models and the contributions of different factors leading to them need to be validated by doctors because they concern patients' health~\cite{dong2017towards,shickel2017deep,cerna2019interpretable}. In cybersecurity, deep learning-based malware detection is aimed at replacing signature-based methods to be practical for antivirus products to recognize and then quarantine/delete malware for computer users. However, a deep learning malware detector that cannot explain why an executable is malicious is unlikely to be completely practical. This is because there are false positives, and malware analysts often need to justify  detection results. The interpretations of a deep learning-based malware detector alleviate malware analysts' efforts of examining them from scratch and creating a knowledge base of malware samples~\cite{cerna2019interpretable}. 

Shicke et al.~\cite{shickel2017deep} make the criticism that deep learning models are hard to interpret, and therefore linear models dominate applied clinical informatics. They also review some attempts to make deep learning models interpretable. For sequential data, Choi et al.~\cite{choi2016retain} propose the interpretable network \textit{RETAIN} to compute the importance of each variable in patients' medical records to their diagnostic predictions. \textit{RETAIN} is composed of two attention-based \textit{RNNs} to form a softmax regression with dynamically computed weights. For image classification, Zeiler et al.~\cite{zeiler2014visualizing} propose Deconvolutional Network (deconvnet) to provide interpretable classification results by revealing which parts of an image are important for its classification. For the classification of vectorial data, logistic/softmax regression and Naive Bayes can interpret how much each feature contributes to a classification result. However, they rely on the feature independence assumption, and thus the interactions of different features cannot be modelled to influence the classification. Inspired by \textit{RETAIN}~\cite{choi2016retain}, we propose a novel multi-layer feed-forward neural network to simulate a logistic regression with a dynamically computed weight of each feature to achieve the same interpretability as logistic regression, while keeping the performance of non-linear models.

\section{Problem Definition}
\label{sec_probdef}

In this section, we define some important concepts, followed by the definition of the research problem.

An executable is a sequence of bytes: 
\begin{eqnarray}
exe = \langle byte_1,byte_2,...\rangle
\end{eqnarray}

The feature set of an executable is extracted by a set of extractors:
\begin{eqnarray}
fea(exe) = \{ext_1(exe),ext_2(exe),...\}
\end{eqnarray}

Except for assembly code, the other extracted features can be represented as a vector. We represent the assembly code as a series of nested sets and sequences.

The assembly code of an executable is a set of assembly functions:
\begin{eqnarray}
code(exe) = \{f_1,f_2,...\}
\end{eqnarray}

An \textit{assembly function} is a set of basic blocks:
\begin{eqnarray}
f = \{b_1,b_2,...\}
\end{eqnarray}

A \textit{basic block} is a sequence of assembly instructions:
\begin{eqnarray}
b=\langle ins_1,ins_2,...\rangle
\end{eqnarray}

An \textit{assembly instruction} is a sequence of one opcode and two operands: 
\begin{eqnarray}
ins = \langle Opcode, Operand1, Operand2\rangle
\end{eqnarray} 
For the uncommon instructions with three operands, the third is ignored. Empty operands are substituted by the special token \texttt{EMPTY}. All opcodes and operands form a set, and each of them is assigned an index number. Thus, one instruction can be abstracted as a sequence of three integers, where each integer represents an index of an opcode or operand. 

\begin{definition}[Malware Detection]\label{defn:problem}
Consider a collection of executables $E$ and a collection of labels $L$ that show the executables in $E$ are benign or malicious. Let $exe$ be an unknown executable that $exe \notin E$. The \textit{malware detection problem} is to build a classification model $M$ based on $E$ and $L$ such that $M$ can be used to determine whether the executable $exe$ is benign or malicious.~\blob
\end{definition}

\begin{figure}
\centering
\includegraphics[width=9cm]{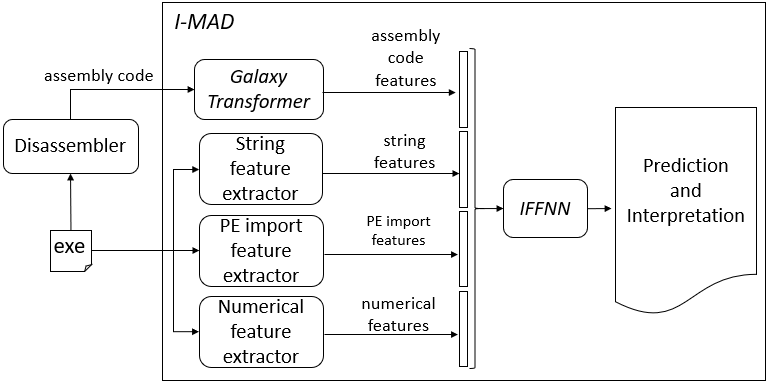}
 \caption{An overview of our \textit{I-MAD} model.}
 \label{fig:overview}
\end{figure}

\section{Methodology}
\label{sec_arc}

Our malware detection model \textit{I-MAD} includes the \textit{Galaxy Transformer} to learn a vector to represent the semantic meaning of the assembly code of an executable and an \textit{interpretable feed-forward neural network} (\textit{IFFNN}) that takes the vector representing the assembly code of a target executable and vectors representing other features as its inputs to produce an interpretable detection result. Figure~\ref{fig:overview} depicts an overview of our malware detection model. In this section, we introduce the \textit{Star Transformer} and describe how we improve it to form the \textit{Star-Plus Transformer} to build the \textit{Galaxy Transformer}. Then, we propose two methods to pre-train different components of the \textit{Galaxy Transformer}. Next, we introduce the other features we use, our novel \textit{IFFNN}, and how we use it to interpret the detection results.

\subsection{Galaxy Transformer}

\subsubsection{Star Transformer}
The \textit{Star Transformer}~\cite{guo2019star} adopts the multi-head attention from the standard \textit{Transformer}:
\begin{align*}
MultiAtt(q,H) &= Concat(head_1,...,head_h)W^O\\
where~head_i&=Attention(qW_i^Q,HW_i^K,HW_i^V)\\
Attention(q_i,K_i,V_i)&=softmax(\frac{q_iK_i^T}{\sqrt{d_{model}}})V_i
\end{align*}
where $K_i=HW_i^K$,$V_i=HW_i^V$, $W_i^Q\in R^{d_{model}\times d_k},W_i^K\in R^{d_{model}\times d_k},W_i^V\in R^{d_{model}\times d_v},W^O\in R^{hd_v\times d_{model}}$ are learnable parameters, $q\in R^{d_k}$ is a query vector, and $H\in R^{n\times d_k}$ is a matrix that contains vector representations of $n$ items to attend to. To compute the self-attention of a sequence $X=\langle x_1,x_2,...,x_n\rangle$, in the \textit{Transformer}, each $x_i$ is a query $q$, and it attends to all items in the sequence, so $H=X$. Thus, its computational complexity is $O(n^2)$.

To reduce the computational complexity, the \textit{Star Transformer} only considers connections between adjacent items and between a relay node and each item, as  shown in Figure \ref{fig:topologies}b. First, for each item $x_i$, a vector $e_i$ is computed as the summation of its non-contextual semantic embedding and its positional encoding in the same way as the \textit{Transformer} does:
\begin{align*}
e_i&=Emb(x_i)=SE(x_i)+PE(i)\\
E&=[e_1;...;e_n]
\end{align*}

Then, the embeddings are fed into a multi-layer neural network to compute the hidden state for each $x_i$. $h_i^t$ represents the hidden state of $x_i$ at layer $t$. $h_i^0$ is initialized as $e_i$. The initial hidden state of the additional relay node is $s^0=\cfrac{1}{n}\sum_n{e_i}$. To compute $h_i^t$, its context matrix $C_i^t$ is formed by the hidden states of itself $h_{i}^{t-1}$ and its adjacent nodes of the previous layer $h_{i-1}^{t-1};h_{i}^{t-1}$, its embedding $e_i$, and the hidden state of the relay node $s^{t-1}$: 
\begin{align*}
C_i^t=[h_{i-1}^{t-1};h_{i}^{t-1};h_{i+1}^{t-1}; e_i;s^{t-1}]
\end{align*}
So, we have $C_i^t\in R^{6d_{model}}$.

At each layer, we have
\begin{align*}
h_i^t&= LayerNorm(ReLU(MultiAtt(h_i^{t-1},C_i^t)))\\
H^t&=[h_1^t;...;h_n^t]\\
s^t&= LayerNorm(ReLU(MultiAtt(s^{t-1},H^t)))
\end{align*}

Thus, the relay node $s^t$ serves as a global information collector. $h_i^t$ collects local information from its adjacency nodes and global information from $s^{t-1}$. The computational complexity to compute all $h_i^t$ is $O(n)$, and to compute $s^t$ it is also $O(n)$. The overall computational complexity is therefore $O(n)$.

To put it all together, we represent a \textit{Star Transformer} Layer as follows:

\begin{align*}
H^{t+1},s^{t+1} = STL^t(H^t,s^t,E)
\end{align*}

The full computation of the \textit{Star Transformer} is as follows:
\begin{align*}
E&=[Emb(x_1);...;Emb(x_n)]\\
H^0&=E,s^0=\cfrac{1}{n}\sum_n{e_i}\\
H^{T},s^{T} &= STL^T(STL^{T-1}(...STL^1(H^0,s^0,E),E),E)
\end{align*}

\subsubsection{Star-Plus Transformer}
As previously shown, the \textit{Star Transformer} can generate a contextual vector representation for each item in a sequence and a vector representation for the whole sequence with 
$O(n)$ computational complexity. We propose the following modifications for better performance.

\begin{enumerate}
\item There is no obvious reason why $e_i$ should be in the context matrix $C_i^t$, so we remove $e_i$ from $C_i^t$, resulting in $C_i^t=[h_{i-1}^{t-1};h_{i}^{t-1};h_{i+1}^{t-1}; s^{t-1}]$. 
\item There was a pointwise feedforward neural network\\ ($FNN = max(0; xW_1 + b_1)W_2 + b_2$) after the multi-head attention computation in the \textit{Transformer}, but it is removed in the \textit{Star Transformer} without an explanation for the rationale. We add it back to compose the information collected by all attention heads and to generate higher-level features for the next layer.
\item A max-pooling on $H^T$ across the top layer mixed with $s^T$ was used as the representation for the whole sequence in the \textit{Star Transformer}. We use only $s^T$ to represent the whole sequence, since it has collected global information of the sequence.
\end{enumerate}

To put it together, we have a \textit{Star-Plus Transformer} layer $H^{t+1},s^{t+1} = SPTL^t(H^t,s^t)$ computed as follows:
\begin{align*}
h_i^t{}'&= LayerNorm(ReLU(MultiAtt(h_i^{t-1},C_i^t)))\\
h_i^t&=LayerNorm(ReLU(FFN(h_i^t{}')))\\
H^t&=[h_1^t;...;h_n^t]\\
s^t{}'&= LayerNorm(ReLU(MultiAtt(s^{t-1},H^t)))\\
s^t&=LayerNorm(ReLU(FFN(s^t{}')))
\end{align*}

\subsection{\textit{Satellite-Planet Transformer} to Understand Basic Blocks}
As we have stated before, a \textit{basic block} is a sequence of assembly instructions: $b=\langle ins_1,ins_2,...\rangle$. The objective of the \textit{Satellite-Planet Transformer} is to learn a vector representation for $b$ using its instructions. To build the \textit{Satellite-Planet Transformer} with the \textit{Star-Plus Transformer}, we modify the input layer of the latter because each instruction is not an atomic item, but a sequence of three items (i.e., an opcode and two operands). Since both the embedding of an instruction $ins_i$ and its positional encoding should have $d_{model}$ dimensions, we make the embeddings of the opcode and operands $d_{model}/3$ dimensions and use the concatenation of them as the embedding of the instruction. It is then added with the positional encoding to form $e_i$. The concatenation of the vector representation of the opcode and operands to form the vector of an instruction was also previously adopted by Ding et al. \cite{ding2019asm2vec}. For the output, we directly use $s^T$, which is the representation of the relay node at the top layer as the semantic meaning representation of the basic block. To train the \textit{Satellite-Planet Transformer} we propose the \textit{Masked Assembly Model} task. 

\begin{definition}[Masked Assembly Model]
Let $(b,ins)$ be a \textit{basic block and assembly instruction pair}. Consider a set of basic block and assembly instruction pairs $B$. For each pair $(b,ins) \in B$, there is one mask instruction $m$ in $b$ that should originally be $ins$. Let $t$ be a target basic block that is not in any pair of $B$, and one of its instructions is replaced by $m$. The \textit{Masked Assembly Model} task is to build a classification model $M$ based on $B$ to predict the original instruction $t$ replaced by $m$.~\blob
\end{definition}

This task is inspired by the \textit{Masked Language Model} task proposed by Devlin et al.~\cite{devlin2018bert}. In that task, the authors mask random words from sentences and use the \textit{Transformer} to predict the masked words based on the contextual words in the sentences. Their method is to feed the output vector of  the \textit{Transformer} corresponding to a masked word to an output softmax over the vocabulary. The prediction requires both global context and local context. The global context means the semantic meaning of the whole sentence except the masked word. The local context means the position of the masked word and its surrounding words that could indicate what ingredient the missing word should be. As the output vector corresponding to the masked word is the only information source for the output layer to make the prediction, it has to capture both global and local context. This does not fit our objective, since the output vector should only contain the semantic meaning of a basic block (i.e., global contextual information). Therefore, we separate the two kinds of information in two vectors: $s^T$ containing the global contextual information and the output vector of the masked instruction $m=[MASK\_OPC, EMPTY, EMPTY]$ containing the local contextual information. We concatenate these two vectors to form one vector and feed it to three feed-forward neural networks with softmax over the whole set of opcodes and operands to predict the opcode and two operands of the original masked instruction. It should be noted that after this training step, we only need to keep the \textit{Satellite-Planet Transformer}, which generates $s^T$, the semantic representation of the entire basic block, because the three feed-forward neural networks to predict the original masked instruction are not needed after the training for the \textit{Masked Assembly Model} task.

\subsection{\textit{Planet-Star Transformer} to Understand Assembly Function}
The \textit{Planet-Star Transformer} is another customized \textit{Star-Plus Transformer} built on top of the \textit{Satellite-Planet Transformer} to learn the vector representation of the semantic meaning of an assembly function $f$ from the set of vectors representing its basic block $\{b_1,b_2,...\}$. As the input is already vectors rather than integers, we abandon the input embedding layer of the \textit{Star-Plus Transformer} that maps integers to embeddings. We directly feed the vectors representing the basic blocks in  positional order to form a sequence to the \textit{Planet-Star Transformer}, which is a \textit{Star-Plus Transformer} without an input layer. We use $s^T$ as the vector representation of the assembly function. To train the \textit{Planet-Star Transformer}, we propose the \textit{Assembly Function Clone Detection} task.

\begin{definition}[Assembly Function Clone Detection]
Let \\$(f_1,f_2)$ be an \textit{assembly function pair}. Let $(f_1,f_2, l)$ be a \textit{labeled assembly function pair} in which the label $l$ indicates whether the two assembly functions $f_1$ and $f_2$ are clones (i.e., semantically equivalent) of each other. Consider a collection of labeled assembly function pairs $F$. Let $p=(f_1,f_2)$ be a new function pair that $p$ is not any function pair in $F$. The \textit{assembly function clone detection task} is to build a classification model $M$ based on $F$ to determine whether the two functions in $p$ are clones of each other.~\blob
\end{definition}

The intuition is that if the vector representations of assembly functions can be used to determine whether two functions are clones of each other, then they contain the semantic meaning of the assembly functions. We train the network to generate similar vectors in cosine measure (i.e., $cos(s^T_{f1},s^T_{f2})$) for real assembly function clone pairs and dissimilar vectors for non-clone pairs. The way we form the function pair dataset is described in Section \ref{sec_exp}.

\subsection{\textit{Star-Galaxy Transformer} to Understand Full Logic of Executable}

Next, we use the \textit{Star-Galaxy Transformer} to learn one vector representing the full logic of an executable based on the representations of all its assembly functions: $\{f_1,f_2,...\}$. Technically, this is similar to learning the representation of an assembly function from the representations of its basic blocks, since both are intended to learn one vector representation from a set of vectors. Therefore, the \textit{Star-Galaxy Transformer} is a duplicate of the \textit{Planet-Star Transformer}. Their difference is that they work at different levels of the hierarchy. The representation of the assembly code of an executable generated by the \textit{Star-Galaxy Transformer} is fed to the \textit{IFFNN} for malware detection without other pre-training tasks proposed for it.

With this, we have completely described how we build the \textit{Galaxy Transformer} with three customized \textit{Star-Plus Transformers} in a hierarchy to compute the vector representation of the assembly code of an executable.

\subsection{Other Features}

When malware is packed, or is polymorphic or metamorphic, the assembly code of its payload is encrypted and not statically accessible. Hence, using only assembly code would fail to identify its malicious purpose. According to the experience of previous works~\cite{anderson2012improving,islam2013classification}, static analysis can still be effective, because the use of the stealthy mechanisms can be captured when analyzed from multiple static feature scopes. Next, we describe the three kinds of features we use and how we improve the way to use them.

\subsubsection{Printable Strings}
According to the literature~\cite{schultz2001data,islam2013classification,dahl2013large,huang2016mtnet}, printable strings are important features, because they include, among others, runtime-linked libraries, functions, and registry keys that are commonly used by malware, system paths, and sometimes the names of user-defined functions. Hence, we extract printable strings from the whole byte sequence of an executable. In our algorithm, a continuous subsequence is a \textit{printable string} if it satisfies three conditions: 1) all of its bytes are ASCII characters, 2) it is terminated with a null symbol, and 3) its length is at least 5 bytes. We count the number of instances of each printable string in the training set and put the strings that appear more than a certain threshold, 1,000 in our case, in the frequent string set. Their frequencies in an executable are used as features. This is not new compared to previous works. The improvement is that we also use the number of printable strings that are not in the frequent string set, i.e., uncommon strings, as a feature, and we use the total number of common printable strings in the executable as another feature. This is based on the intuition that encrypted malware has more uncommon printable strings and benign software has more common strings.

\subsubsection{PE Imports}
PE Imports are dynamically linked libraries and functions shown in the import address table of PE headers. The imports of an executable often illustrate its behaviors, e.g., modify the registry or hook a procedure~\cite{schultz2001data,saxe2015deep}. The total number of imports show whether the executable is hiding its potential behaviors, because abnormally few imports indicate that runtime linking is largely used or most of its imports are hidden in encrypted data. Therefore, we compute these features in the same way as we compute the printable string features.

\subsubsection{PE Header Numerical Features}
There are many numerical fields in PE headers that contain information that could form different patterns among malware and benign software (benignware)~\cite{baldangombo2013static,saxe2015deep}. Hence, we also use these values as features.

We concatenate the vector representing the full logic of an executable $v_{code}$, printable string feature vector $v_{str}$, PE header numerical feature vector $v_{num}$, and PE import feature vector $v_{imp}$ to form a vector representation of the executable from multiple scopes $v=[v_{code},v_{str},v_{num},v_{imp}]$.

\subsection{Interpretable Feed-Forward Neural Network}

\begin{figure}
\centering
\includegraphics[width=8cm]{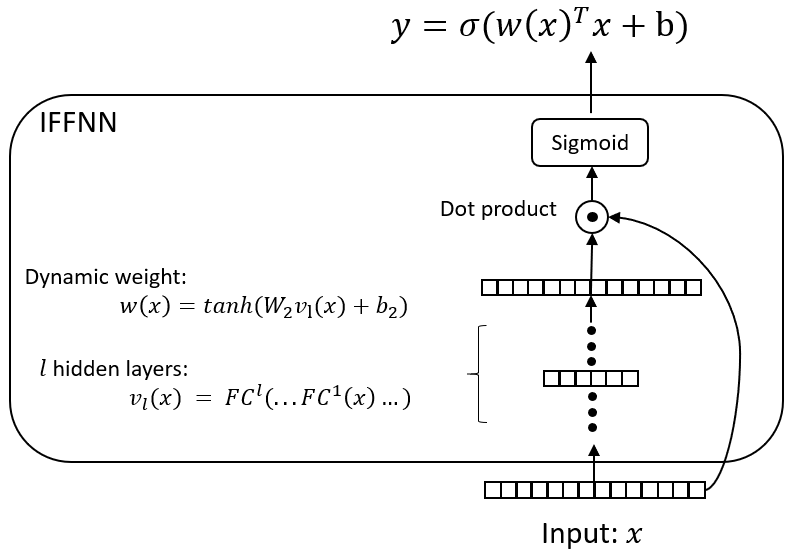}
 \caption{Our proposed \textit{IFFNN}.}
 \label{fig:iffnn}
\end{figure}

Interpretability is an important quality of a machine learning model for malware detection. Inspired by the work of Choi et al.~\cite{choi2016retain}, which uses two attention-based \textit{RNN} networks to form a softmax regression model with dynamic weights, we propose a novel interpretable feed-forward neural network (\textit{IFFNN}) to form a "dynamic logistic regression" model. Figure~\ref{fig:iffnn} illustrates its architecture.

Let $x\in \mathbb{R}^m$ be a feature vector representing a sample. We first feed it to $l$ fully-connected hidden layers:
\begin{align}
v_{l}(x) &= FC^{l}(...FC^{1}(x)...)\\
where~FC^{i}(v_{i-1}(x))&=tanh(W_1^iv_{i-1}(x)+b_1^i)
\end{align}
where $W_1^i\in\mathbb{R}^{d_h^i\times d_h^{i-1}}$, $b_1^i\in\mathbb{R}^{d_{h}^i}$, and $v_{l}(x)\in\mathbb{R}^{d_{h}^l}$. Then, we apply another normal fully-connected layer of which the output vector has the same dimension as $x$:
\begin{equation}
w(x) = W_2v_{l}(x)+b_2
\end{equation}
where $W_2\in\mathbb{R}^{m\times d_{h}^l}$, $b_2\in\mathbb{R}^{m}$, and $w(x)\in\mathbb{R}^{m}$. $w(x)$ serves as a weight vector for each feature in $x$. The final confidence that the input sample is positive (in malware detection, positive means malicious) is calculated as follows:
\begin{align}
y = IFFNN(x) &= \sigma(w(x)^Tx+b)\\
where~\sigma(z)&=\cfrac{1}{1+e^{-z}}
\end{align}
where $b\in \mathbb{R}$ is a bias term. This is similar to a logistic regression (i.e., $y=\sigma(w^Tx+b)$, where $w$ is a parameter vector), and the difference is that our weight vector $w(x)$ is dynamically computed based on $x$ rather than static parameters.

As can be seen, the \textit{IFFNN} has the same interpretability (a.k.a., intelligibility~\cite{lou2012intelligible}) as logistic regression because of its decomposability and algorithmic transparency~\cite{lipton2018mythos} when the features have certain meanings. Since the weight of each feature is dynamically computed by a multi-layer neural network, the feature interactions are still modelled and the weight of each feature to the prediction is contextualized. Thus, it has the modelling power of a non-linear model. 

The \textit{IFFNN} can be used for any binomial classification task and can be plainly generalized to a "dynamic softmax regression" model for multinomial classification, as long as the sample can be represented as a vector with fixed dimension.

We feed $v$, the feature vector from multiple scopes of an executable, to the proposed \textit{IFFNN} to get the confidence $y$ that it is malicious: $y = IFFNN(v)$ and interpret the result. Thus, we complete the full network of the top-level model. 

\subsection{Attribution}

For logistic regression: $y=\sigma(w^Tx+b)=\sigma(w_1x_1+w_2x_2+...+w_mx_m+b)$, where $x=(x_1,x_2,...,x_m)$ and $w=(w_1,w_2,...,w_m)$, the attribution is simple. Whether feature $x_j$ makes the sample positive depends on the sign of $w_jx_j$. If $w_jx_j > 0$, $x_j$ makes it positive, and vice versa. The degree of the impact of $x_j$ on $y$ depends on $|w_jx_j|$: a large $|w_jx_j|$ implies a large impact of $x_j$. If the model predicts a sample to be positive, the most influential factor that leads to the result is $max_j{w_jx_j}$. If the model predicts a sample to be negative, the most influential factor that leads to the result is $min_j{w_jx_j}$.

The same idea is applicable to our \textit{IFFNN} for malware detection. Its top layer is logistic regression with dynamically computed weight: $y = \sigma(w(v)^Tv)=\sigma(w(v)_1v_1+w(v)_2v_2+...+w(v)_mv_m)$. If $|w(v)_jv_j|$ is large and $w(v)_jv_j>0$, feature $v_j$ has a large impact on the prediction of malicious. If $|w(v)_jv_j|$ is large and $w(v)_jv_j<0$, feature $v_j$ has a large impact on the prediction of benign. For printable string features, PE imports, and PE header numerical features, each dimension of their vector representations corresponds to a specific feature. The features can be the frequency of a certain string, whether a certain DLL is imported, the value of a certain numerical field, etc. By checking its $w(v)_jv_j$, we know whether it makes the executable more likely malicious or benign. For the vector representing the full logic of an executable: $v_{code}$, each of its dimensions has no specific meaning, but we can see the impact of the full logic of the executable by computing the summation of the impact of each dimension of its vector: $\sum_{j\in v_{code}} w_{code,j}v_{code,j}$. If it is positive, from the perspective of the assembly code, the executable is more likely malicious, and vice versa. 

As $v_{code}$ is computed by our \textit{Star-Galaxy Transformer} network, the attention weights of the assembly functions to the relay node at the top layer indicate the importance of each assembly function. We compute the summed attention weights of each assembly function over all heads to the relay node to determine which assembly functions are the main factors that influence the classification results.

\subsection{Model Training}
To train the \textit{Satellite-Planet Transformer}, the objective function is the cross entropy loss of the prediction on the masked opcode and operands against the real opcode and operands. To train the \textit{Planet-Star Transformer} and simultaneously fine-tune the \textit{Satellite-Planet Transformer}, the objective function is the mean squared error between the computed cosine similarity between two assembly functions and the gold standard (i.e., 1 for clone function pairs, and -1 for non-clone function pairs). To train the full top-level network including the \textit{IFFNN} and the \textit{Star-Galaxy Transformer}, the objective function is the cross entropy loss of the prediction against the real label. To ensure that the \textit{Star-Galaxy Transformer} gets sufficient training, we first train it without concatenating any other feature, i.e., feed $v_{code}$ instead of $v$ to the \textit{IFFNN} ($y=IFFNN(v_{code})$), and train it for malware detection. This is in fact the pre-training of the \textit{Star-Galaxy Transformer}. Then, we concatenate $v_{code}$ with other features to feed it to the \textit{IFFNN} ($y=IFFNN(v)$), and train it the same way for malware detection. The \textit{Satellite-Planet Transformer} and \textit{Planet-Star Transformer} networks are not fine-tuned when we train the top-level network. For all the training objectives, we use Adam~\cite{kingma2014adam} with the initial learning rate $1e-4$. We use early stopping with the validation set to avoid overfitting \cite{caruana2001overfitting}.

\section{Experiments}
\label{sec_exp}

The objectives of our experiments are to 1) evaluate the performance of \textit{I-MAD} for malware detection, 2) compare \textit{I-MAD} to other state-of-the-art static malware detection solutions, and 3) demonstrate the interpretability of \textit{I-MAD}. 

We train and evaluate the models on a server with two Xeon E5-2697 CPUs, 384 GB of memory, and four Nvidia Titan XP graphics cards. We use PyTorch~\cite{paszke2017automatic} to implement our model. We use the "pefile"\footnote{https://github.com/erocarrera/pefile} library to extract numerical features from PE headers.

\subsection{Datasets and Pre-training}

For the two pre-training tasks, we compile several open source projects that are compatible with GCC and/or LLVM. We choose these two compilers because they are the most appropriate options to provide different compilation options to generate semantically equal but literally different assembly functions. GCC compiler provides four different optimization levels (i.e., O0, O1, O2, and O3) to compile projects. We compile \textit{busybox}, \textit{coreutils}, \textit{libcurl}, \textit{libgmp}, \textit{libtomcrypt}, \textit{libz}, \textit{magick}, \textit{openssl}, \textit{puttygen}, and \textit{sqlite3} with GCC at all four optimization levels. Thus, for every assembly function in those projects we have four semantically equivalent versions. O-LLVM\footnote{https://github.com/obfuscator-llvm/obfuscator/wiki} is an obfuscator of the LLVM compiler that provides control flow flattening, instruction substitution, and bogus control flow obfuscation mechanisms. We use O-LLVM to compile \textit{libcrypto}, \textit{libgmp}, \textit{libMagickCore}, and \textit{libtomcrypt} with five different settings: no obfuscation, each of the three obfuscation mechanisms, and all three mechanisms. Thus, we have five versions of their every function. We use IDA Pro\footnote{https://www.hex-rays.com/products/ida/}, a commercial disassembler, to disassemble our compiled executables and acquire the assembly functions. 

We use basic blocks of lengths between 5 to 250 instructions to form our \textit{Masked Assembly Model} dataset; these blocks are within the typical length range of blocks that provide enough context and are not too long to harm training efficiency. As a result, this dataset contains 38,427,440 basic blocks. We use all of them for training and none for testing as the purpose of the dataset is to train the \textit{Satellite-Planet Transformer} to understand assembly code, and the accuracy of this task is uninformative.

We use the semantically equivalent but literally different functions we compiled to form real function clone pairs. We randomly pair the same number of functions to be non-clone function pairs to create the dataset for the \textit{Assembly Function Clone Detection} task. We limit the maximum number of instructions per basic block to be 50 and the maximum number of blocks per function to be 50 in this dataset, so that the memory of our graphics cards can hold the data flowing in the bottom two-level networks. There are 213,656 function pairs in the training set, 26,898 functions in the validation set, and 26,746 functions in the test set. Our bottom two-level networks get a classification accuracy of 91.5\% on the test set. This means that the assembly function representations it computes and the representations of basic blocks that are fed to it indeed capture the semantic meanings of assembly code. We do not elaborate on the experiments for this task since it is not the objective task, but rather a task to pre-train the \textit{Planet-Star Transformer} and fine-tune the \textit{Satellite-Planet Transformer}.
\begin{table}[h]
\caption{Top 10 majority malware families of the dataset.}
\label{tab:dataset}
\begin{center}
\begin{tabular}{|c|c|c|}
\hline
Malware Family&Number&Percentage\\
\hline
Fareit&9,436&8.2\%\\
Zbot&6,433&5.6\%\\
Emotet&6,343&5.5\%\\
Gandcrab&4,120&3.6\%\\
Mepaow&4,055&3.5\%\\
CobaltStrike&3,151&2.7\%\\
Allaple&2,081&1.8\%\\
Ursnif&1,552&1.3\%\\
Autoit&1,017&1.0\%\\
NaKocTb&794&0.7\%\\
\hline
Total&38,982&33.9\%\\
\hline
\end{tabular}\vspace{-5mm}
\end{center}
\end{table}
For malware detection we collected a dataset containing 115,000 benign and 115,000 malicious executables. There is no redundancy in the dataset. Following the literature~\cite{schultz2001data,kolter2004learning,raff2017malware}, the benign executables are the .exe and .dll files from the installation paths of software programs. The malicious executables are collected from \emph{MalShare} and \emph{VirusShare}. The top 10 major malware families of the dataset are presented in Table~\ref{tab:dataset}. They are obtained with ClamAV~\footnote{https://www.clamav.net/}. The top 10 known packers that are applied on the malware samples are shown in Table~\ref{tab:packing}. The usage of packers is acquired with Yara Rules~\footnote{https://github.com/Yara-Rules/rules}. The way we split the dataset into training set, validation set, and test set is introduced in Subsection~\ref{subsec:expset}.
\begin{table}[h]
\caption{Top 10 packers used in the malware dataset.}
\label{tab:packing}
\begin{center}
\begin{tabular}{|c|c|c|}
\hline
Packer&Number&Percentage\\
\hline
UPX&7,776&6.7\%\\
BobSoft Mini Delphi&5,262&4.5\%\\
ASProtect&1,826&1.59\%\\
ASPack&1,780&1.55\%\\
PECompact&586&0.51\%\\
Armadillo&369&0.32\%\\
D1S1G&155&0.14\%\\
WinrarSFX&124&0.11\%\\
MoleBox&69&0.06\%\\
WinZipSFX&38&0.03\%\\
\hline
Total&17,985&15.6\%\\
\hline
\end{tabular}\vspace{-5mm}
\end{center}
\end{table}
\subsection{Models for Comparison}

\begin{table*}[h]
\caption{Results of k-fold cross-validation experiment. It includes the p-values (pv) of t-test for F1 and accuracy between \textit{I-MAD} (ST+) and other models.}
\label{tab:det_result}
\begin{center}
\begin{tabular}{|c|c|c|c|c|c|c|}
\hline
Model&P&R&F1&pv (F1)&Acc &pv (Acc)\\
\hline
\textit{Mosk2008OB}&96.1&95.8&95.9&3.3e-13&95.9&1.6e-20\\
\textit{Bald2013Meta}&96.5&95.9&\textit{\textbf{96.2}}&1.1e-13&\textit{\textbf{96.2}}&6.7e-20\\
\textit{Saxe2015Deep}&95.2&96.1&95.7&4.0e-14&95.6&4.5e-21\\
\textit{Raff2017MalC}&95.9&96.3&96.1&5.6e-15&96.1&4.0e-20\\
\textit{Krcal2018Conv}&93.2&93.2&93.2&1.7e-15&93.2&1.0e-23\\
\textit{Mour2019CNN}&72.6&71.5&72.0&2.3e-26&71.8&1.5e-30\\\hline
\textit{SVM} (same features)&96.1&96.4&\textit{\textbf{96.2}}&3.7e-13&\textit{\textbf{96.2}}&5.6e-20    \\\hline
\textit{I-MAD} (no code)&96.5&96.6&96.5&9.8e-13&96.5&4.0e-19\\
\textit{I-MAD} (ST)&97.0&97.9&97.3&5.0e-3&97.2&4.7e-6\\
\textit{I-MAD} (ST+)&97.5&97.9&\textbf{97.7}&N/A&\textbf{97.7}&N/A\\\hline
\end{tabular}\vspace{-5mm}
\end{center}
\end{table*}
We compare our \textit{I-MAD} model to several state-of-the-art static malware detection models.
\begin{itemize}
\item  \textbf{Mosk2008OB} Moskovitch et al.~\cite{moskovitch2008unknown} propose to use TF or TF-IDF of opcode bi-grams as features and use document frequency (DF), information gain ratio, or Fisher score as the criteria for feature selection. They apply Artificial Neural Networks, Decision Trees, Naïve Bayes, Boosted Decision Trees, and Boosted Naïve Bayes as their malware detection models.

\item  \textbf{Bald2013Meta} Baldangombo et al.~\cite{baldangombo2013static} propose to extract multiple raw features from PE headers and use information gain and calling frequencies for feature selection and PCA for dimension reduction. They apply SVM, J48, and Naïve Bayes as their malware detection models.

\item  \textbf{Saxe2015Deep} Saxe et al.~\cite{saxe2015deep} propose a sophisticated deep learning model that works on four different features: byte/entropy histogram features, PE import features, string 2D histogram features, and PE metadata numerical features. We tried to follow the exact features they extract when we implement it, but they do not provide the exact metadata numerical fields they use, so we just use the same numerical fields of PE headers used in our model as part of their input.

\item  \textbf{Raff2017MalC} Raff et al.~\cite{raff2017malware} treat an executable as a sequence of bytes and apply a gated 1D convolutional neural network (CNN) to classify an executable. The network includes an embedding layer, two convolutional layers with large filters and strides, a global max-pooling layer, and two fully-connected layers. The output of one convolutional layer serves as the gate of the other.

\item  \textbf{Krcal2018Conv} Following Raff et al.~\cite{raff2017malware}, Krcal et al.~\cite{krvcal2018deep} treat an executable as a sequence of bytes and apply a CNN for malware detection, but their CNN is deeper and has smaller filters. There are four convolutional layers and four fully connected layers. Instead of a global max-pooling layer, they use a global mean-pooling layer after the convolutional layers.

\item  \textbf{Mour2019CNN} Mourtaji et al.~\cite{mourtaji2019intelligent} convert malware binaries to grayscale images and apply a 2D CNN on malware images for malware classification. 
\end{itemize}

For the papers in which the authors describe multiple ways to select features and/or apply multiple machine learning models~(\cite{moskovitch2008unknown,baldangombo2013static}), we try with all possible settings and report the best results that their methods can achieve to compare with our model.

As the ablation study, we also compare our full model "\textit{I-MAD} (ST+)" with "\textit{I-MAD} (no code)" and "\textit{I-MAD} (ST)". "\textit{I-MAD} (no code)" is our model without using assembly code. These comparisons can show the effectiveness of modeling assembly code with \textit{Galaxy Transformer}. "\textit{I-MAD} (ST)" is to build the \textit{Galaxy Transformer} with the original \textit{Star Transformer}, rather than the \textit{Star-Plus Transformer}, to show the effectiveness of our modifications.

We also compare our model with an SVM model that uses the same features as \textit{I-MAD} except for assembly code, since it is not a vectorial feature. We consider linear, polynomial, and RBF kernels and use grid search for tuning hyper-parameters. Comparing this baseline with \textit{I-MAD} (no code), we can separately show the effectiveness of the feature set and our model.

\subsection{Experiment Settings}
\label{subsec:expset}
We evaluate the models under two different experiment settings. The main evaluation metric is accuracy (Acc), but we also evaluate the models with precision (P), recall (R), and F1.
\begin{itemize}
\item  \textbf{K-Fold Cross-Validation} We first evaluate our model and others with k-fold cross-validation where $k=5$. The original dataset is randomly split into 5 even subsets. Each subset takes a turn to be chosen as the test set. Another subset takes a turn to be chosen as the validation set. The other 3 subsets form the training set. Thus, we have $^5P_2=20$ different experiment groups. Each group contains 138,000 samples in the training set, 46,000 in the validation set, and 46,000 in the test set. We acquire the experiment results of the 20 groups and report the averages.

\item  \textbf{Time Split Evaluation} In addition to cross-validation evaluation, we also evaluate the models in a more challenging and realistic scenario. In real life, a malware detection system is expected to detect new malware with its knowledge of known malware. To evaluate this ability of the models, we follow Saxe et al.~\cite{saxe2015deep} to perform a time split experiment. We use the executables compiled before 2015 to form the training and validation set, and those compiled after 2017 to form the test set. We exclude samples with a compilation time before 2000 or after 2020, either because the compilation dates are fake or the samples are outdated. There are 106,000 samples in the training set, 20,000 in the validation set, and 40,000 in the test set. We run each model with different initialization and random seeds 5 times and report the averages of the results.
\end{itemize}

\begin{table*}[h]
\caption{Results of time split experiment. It includes the p-values of t-test for F1 and accuracy between \textit{I-MAD} (ST+) and other models.}
\label{tab:time_split}
\begin{center}
\begin{tabular}{|c|c|c|c|c|c|c|}
\hline
Model&P&R&F1&pv (F1)&Acc &pv (Acc)\\
\hline
\textit{Mosk2008OB}&88.6&88.6&88.6&1.2e-15&88.6&3.4e-22\\
\textit{Bald2013Meta}&88.3&88.1&88.2&1.6e-17&88.2&1.2e-22\\
\textit{Saxe2015Deep}&87.4&87.7&87.5&1.4e-17&87.5&2.6e-23\\
\textit{Raff2017MalC}&88.5&89.0&\textit{\textbf{88.7}}&1.2e-16&\textit{\textbf{88.7}}&4.5e-22\\
\textit{Krcal2018Conv}&84.2&83.2&83.7&3.1e-18&83.8&1.2e-27\\
\textit{Mour2019CNN}&57.0&56.6&56.8&4.3e-31&56.9&7.1e-34\\\hline
\textit{SVM (same features)}        &89.2&88.8&\textit{\textbf{89.0}}&7.3e-15&\textit{\textbf{89.0}}&6.1e-22\\\hline
\textit{I-MAD} (no code)&89.4&89.6&89.5&3.2e-15&89.5&6.7e-21\\
\textit{I-MAD} (ST)&91.1&91.1&91.1&8.6e-11&91.1&2.6e-15\\
\textit{I-MAD} (ST+)&91.4&91.6&\textbf{91.5}&N/A&\textbf{91.5}&N/A\\\hline
\end{tabular}\vspace{-5mm}
\end{center}
\end{table*}

\subsection{Results}

The results of the k-fold cross-validation and the time split experiments are shown in Table~\ref{tab:det_result} and Table~\ref{tab:time_split}, respectively.

The full version of \textit{I-MAD} achieves statistically significantly better accuracy and F1 than the other models in all experiments, as the p-values in t-test are much smaller than $0.01$. The improvements of our model on accuracy and F1 are larger in the time split experiments than in the cross-validation experiment. Even though we make sure there is no redundancy in the dataset, some pieces of malware could be extensively similar to each other if they are from the same family and compiled with slightly different modifications. Also, their compilation time is usually close to each other. In the time split experiment, the executables in the test set are compiled at least 2 years later than any executable in the training and validation sets. This is a more difficult setting that can be reflected in the pervasively lower accuracy in the time split setting than in the cross-validation setting. Thus, the significantly larger improvement of our detection model over other models in the time split experiment indicates that it has better abilities to learn robust and consistent patterns from old samples that can be generalized to classify new samples.

It is clear that with modelling assembly code with the \textit{Galaxy Transformer}, \textit{I-MAD} achieves much better results than it does without modelling the assembly code. This shows that modelling assembly code with our \textit{Galaxy Transformer} helps in differentiating malicious and benign executables. We can also see that the \textit{Galaxy Transformer} built with \textit{Star-Plus Transformer} (\textit{I-MAD} (ST+)) is more effective than the one built with the original \textit{Star Transformer} (\textit{I-MAD} (ST)). This confirms that our modifications are useful.

SVM with the same features as \textit{I-MAD} except for assembly code, achieves accuracy similar to other best baseline models in the cross-validation experiment, and it achieves better accuracy than other baseline methods in the time split experiment, while worse than \textit{I-MAD} (no code). This shows that the feature set we propose is effective, and our \textit{IFFNN} has advantages in classification performance on the same feature set.

That being said, other models, except \textit{Mour2019CNN}, also achieve reasonably good results in all experiments. However, none of the models consistently achieves the second-best performance in both experiment settings. Even though \textit{Saxe2015Deep} uses features from multiple scopes, they do not show better results than \textit{Bald2013Meta} and \textit{Mosk2008OB}. The lack of any mechanism to understand assembly code is an obvious reason, as modelling assembly code in our model improves the performance. Our improved way of representing printable string and PE import features, combined with our \textit{IFFNN}, is the other reason. This is validated in the next subsection.

\textit{Mour2019CNN} performs much worse than other models, even though we tried alternative hyper-parameter values in addition to the values the authors provided. One reason is that the way it represents an executable as an image is not sophisticated; even a small offset change in an executable would result in totally different textures in its image. In addition, we also observe overfitting, as its accuracy on the training set achieves 89.2\%, while on the test set it is 71.8\%. Even though our model is also a deep learning model, it does not suffer from the overfitting problem because we use two pre-training tasks to adequately train the \textit{Satellite-Planet Transformer} and \textit{Planet-Star Transformer} with the rich information embedded in assembly code. In contrast, \textit{Mour2019CNN} can only be trained with the labels of executables, which is insufficient.

\subsection{Interpretability}

\subsubsection{Case Study}
Table~\ref{tab:example} shows how our model interprets the detection result of a sample. The primary factors that lead to the prediction of 05c199.exe to be malicious and the main assembly functions related to the prediction are given. It can be seen that the assembly code of the target executable is the primary reason. The two assembly functions that contribute the most to the prediction set the program to sleep for a certain time and then download and run an embedded executable from a remote address.

\subsubsection{Qualitative Analysis}
To better understand the impacts of the features we use, Table~\ref{tab:sta_att} shows the ten most frequent main factors leading to the prediction of a sample to be malware or benign.

\begin{table}[h]
\caption{Most frequent main factors leading to the predictions of the malicious or benign class.}
\label{tab:sta_att}
\begin{center}
\begin{tabular}{|c|}
\hline
Main factors leading to the prediction of the malicious class\\
\hline
Assembly code\\
Total number of PE imports\\
Number of uncommon strings\\
The frequency of the string "Password"\\
The import of \textit{KERNEL32.dll}\\
Total number of strings\\
The import of \textit{WriteFile}\\
The frequency of the string "\textbackslash x02\textbackslash x02GetLastError"\\
Subsystem\\
Maximum entropy of sections\\
\hline
Main factors leading to the prediction of the benign class\\
\hline
Total number of strings\\
Number of uncommon strings\\
The import of \textit{LCMapStringW}\\
Total number of PE imports\\
Assembly code\\
Maximum entropy of sections\\
The frequency of the string "\textit{\textbackslash r\textbackslash x01\textbackslash x01\textbackslash x01\textbackslash x05}"\\
The frequency of the string "\textit{\textbackslash r\textbackslash x01\textbackslash x01\textbackslash x05\textbackslash x05}"\\
The import of \textit{initterm}\\
Mean entropy of sections\\
\hline
\end{tabular}
\end{center}
\end{table}

\paragraph{Main factors for both classes} The assembly code of an executable is one of the most frequent factors influencing the prediction of an executable to be malicious or benign. This means that the vector representing the semantic meaning of assembly code computed by our \textit{Galaxy Transformer} is very effective for malware detection. We randomly examine the assembly functions of some malware that acquire the largest attention by the relay node at the top layer of the \textit{Star-Galaxy Transformer}. Many of them concern malicious behaviors, such as installing itself into some registry, hijacking some common legitimate DLLs, and injecting itself into another process. We find that there are statistical differences between the two classes in the mean values of total number of strings, number of uncommon strings, total number of PE imports, and maximum entropy of the sections. To be more specific, on average there are less common strings, more uncommon strings, less PE imports, and higher entropy among malware. The fact that these features could be main factors for both classes also shows the superiority of our \textit{IFFNN} over logistic regression: as the number of uncommon strings and the number of PE imports always have non-negative values, when each of them serves as a main factor leading to the prediction of the malicious class, its weight is positive (i.e., $w(v)_jv_j>0\&v_j>0\Rightarrow w(v)_j>0$), and when it serves as a main factor leading to the prediction of the benign class, its weight is negative (i.e., $w(v)_jv_j<0\&v_j>0\Rightarrow w(v)_j<0$). This cannot be achieved by logistic regression because when it is trained, the weight for each feature is determined and stays static, irrelevant of the input samples. However, the weight of each feature in \textit{IFFNN} is dynamically computed based on the whole context, i.e., the vector representing all features. 

The explanation from the perspective of statistics is as follows. All supervised machine learning classification models work by identifying the correlation between a feature and a class. Logistic regression can only learn the independent correlation between a feature and a class, without considering the correlation between features; therefore, it is linear and the weight for each feature is static. \textit{IFFNN} learns the correlation between a feature and a class in a context considering the correlations between different features. 

\paragraph{Main factors for malicious class} The import of "KERNEL32.dll" is a main factor for the prediction of malicious class because malware relies heavily on a large number of core APIs in it to manipulate memory and the file system. The "WriteFile" function is also a main factor because malware such as ransomware and worms uses it to write content to the file system. The string of "Password" is another main factor that more frequently appears in malware created for credential theft purposes. Malware often uses mutex for different reasons. For example, it can be used as a locking mechanism to serialize access to a resource on the system or to avoid more than one instance of itself running. "GetLastError" is used to determine whether a mutex already exists. This is the reason why the frequency of string "\textbackslash x02\textbackslash x02GetLastError" is a main factor leading to the prediction of malware.

\paragraph{Main factors for benign class} "LCMapStringW" is often used by benign software to convert all characters of strings to upper/lower case, which is a feature much less used in malware. "initterm" is used by core libraries to initialize a function pointer table and does not need to be imported by software programs, and therefore it is an indicator of some benign libraries. "\textit{\textbackslash r\textbackslash x01\textbackslash x01\textbackslash x01\textbackslash x05}" and "\textit{\textbackslash r\textbackslash x01\textbackslash x01\textbackslash x05\textbackslash x05}" are two strings that appear 1.8 and 3.6 times respectively more frequently among benign executables than malicious executables.

\subsubsection{Quantitative Analysis}
We also use a quantitative measure to analyze the interpretation of \textit{I-MAD}. We compute the \textit{Gini importance (GI)} and \textit{information gain (IG)} of the features, and then rank them based on those criteria. We then rank the features by the frequencies that they serve as the main factors for the predictions. Features serving as main factors more frequently should be relatively important features for malware detection. It should be noted that even though the importance ranked this way is relevant to the rank by \textit{Gini importance} or \textit{information gain}, they are not supposed to be equivalent. Even if the attribution mechanism of \textit{I-MAD} gives a perfect interpretation, the feature importance rank based on that would still be different from the rank by \textit{Gini importance} or \textit{information gain}. 

Table~\ref{tab:spearman} shows the Spearman's Rank Correlation Coefficient between the rank given by \textit{I-MAD}, \textit{Gini importance}, and \textit{information gain}. It can be seen that the Spearman's Rank Correlation Coefficient between the rank given by \textit{I-MAD} and those given by \textit{Gini importance} and \textit{information gain} are 0.59 and 0.55, respectively. This shows a strong correlation between them. The correlation coefficient between the rank by information gain and by Gini importance is only 0.72, even though they are often used for the exact same purpose: feature selection. The result means that the \textit{IFFNN} in \textit{I-MAD} frequently uses features that have high information gain or Gini importance as its main classification factors.

\begin{table}[h]
\caption{The Spearman's Rank Correlation Coefficient between the feature importance rank given by \textit{I-MAD}, \textit{Gini importance}, and \textit{information gain}.}
\label{tab:spearman}
\begin{center}
\begin{tabular}{|c|c|c|c|}
\hline
&IG&GI&\textit{I-MAD}\\\hline
IG&1.0&0.72&0.59\\\hline
GI&0.72&1.0&0.55\\\hline
\end{tabular}
\end{center}
\end{table}
\begin{table}[h]
\caption{Efficiency of each model in terms of number of samples classified per second. The time consumption for feature extraction is not included.}
\label{tab:efficiency}
\begin{center}
\begin{tabular}{|c|c|}
\hline
Model&n samples per second\\
\hline
\textit{Mosk2008OB}&32,152\\
\textit{Bald2013Meta}&127,988\\
\textit{Saxe2015Deep}&142,711\\
\textit{Raff2017MalC}&86\\
\textit{Krcal2018Conv}&142\\
\textit{Mour2019CNN}&391\\\hline
\textit{SVM (same features)}&58\\\hline
\textit{I-MAD} (no code)&28,197\\
\textit{I-MAD} (ST)&15,355\\
\textit{I-MAD} (ST+)&15,239\\\hline
\end{tabular}\vspace{-5mm}
\end{center}
\end{table}
\subsection{Efficiency Study}
The efficiency of \textit{I-MAD} and all models for comparison measured by the number of samples classified per second is presented in Table~\ref{tab:efficiency}. 

Among all models, the efficiency of \textit{I-MAD} is moderate. And \textit{I-MAD} is the second most efficient deep learning model. \textit{Saxe2015Deep} is the most efficient because the dimension of its feature vector is only 1024, and the network is very small. \textit{Raff2017MalC} and \textit{Krcal2018Conv} are slow because they rely on whole byte sequences, and they are very computationally expensive. With our Titan Xp graphics cards their batch sizes could be around 32 and 128 at most, respectively. The batch size for \textit{Mour2019CNN} depends on the number of bytes in the samples; in extreme cases we need to run the model on the CPU because the graphics card memory cannot hold the computation for even one large executable. For \textit{I-MAD} (ST)/(ST+), the batch size could be at least 512 for most samples. As the representation of assembly code is computed at three levels (i.e., basic block, function, and executable), the memory for the lower level computation is released and reused after the representation is computed. For \textit{I-MAD} (no code), the batch size could be 5,120. It is worth the extra  computational cost to model assembly code because the benefit of it in classification performance is significant. SVM with the same features as \textit{I-MAD} also has very low efficiency because its computational complexity is linear with the dimension of feature vector and the number of support vectors, which are large when the dataset is complex. In our experiments, there are always more than 43,000 support vectors, and the dimension of feature vectors is more than 2,700.

\section{Limitations and Future Work}
 \label{sec_lim}
Adversarial attack and defense are closely related topics to classification problems such as image classification~\cite{goodfellow2014explaining} and malware detection~\cite{anderson2018learning}. Interpretability is a double-edged sword considering adversarial attacks in white-box settings, where adversaries have full access to the \textit{I-MAD} model and  can use the interpretations to craft adversarial samples more easily than by using an uninterpretable model. 

Evasion techniques (e.g., adversarial attacks) are extensively applied in wild malware, as is the case of our dataset. Following previous experience~\cite{anderson2012improving,islam2013classification}, we counter the evasion techniques by detecting malware from the views of multiple disparate feature sets. Also, since adversarial samples are already in our dataset, adversarial training is automatically performed to defend against the attacks~\cite{goodfellow2014explaining}. Usually, adversarial attack and defense are discussed in a different research paper than the one proposing a novel classification model. One of our future work directions is to further investigate adversarial attack and defense on malware detection.

\section{Conclusion}
 \label{sec_con}

In this paper, we present \textit{I-MAD}, a novel deep learning model for static malware detection that is based on the understanding of assembly code. In addition to its excellent detection performance, it can also provide interpretation for its detection results, which can be examined by malware analysts. Therefore, in addition to malware detection, it can also help malware analysts locate malicious payloads and find consistent patterns in malware samples. The proposed \textit{IFFNN} has values that can be applied in interpretable classification for other tasks as well.

\printcredits

\section*{Acknowledgments}
This research is supported by Defence Research and Development Canada (contract no. W7701-176483/001/QCL), NSERC Discovery Grants (RGPIN-2018-03872), and Canada Research Chairs Program (950-232791). The Titan Xp used for this research was donated by the NVIDIA Corporation.

\bibliographystyle{bibstyle}
\bibliography{main}

\bio{}
\textbf{Miles Q. Li} is a Ph.D. candidate in the School of Computer Science, McGill University, Montreal, Canada. He received his B.Sc and M.Sc from Peking University. His research interests include data mining for cybersecurity, deep learning, and natural language processing.
\endbio

\bio{} 
\textbf{Benjamin C. M. Fung} is a Canada Research Chair in Data Mining for Cybersecurity, a Full Professor with the School of Information Studies, and an Associate Member with the School of Computer Science at McGill University in Canada. He received a Ph.D. degree in computing science from Simon Fraser University in Canada in 2007. He has over 130 refereed publications that span the research forums of data mining, privacy protection, cybersecurity, services computing, and building engineering. His data mining works in crime investigation and authorship analysis have been reported by media worldwide. Prof. Fung is a licensed Professional Engineer of software engineering in Ontario, Canada.
\endbio

\bio{}
\textbf{Philippe Charland} is a Defence Scientist at Defence Research and Development Canada – Valcartier Research Centre, in the Mission Critical Cyber Security Section. His research interests encompass software reverse engineering and computer forensics. As a member of the Systems Vulnerabilities and Lethality Group, his research focuses on binary-level program comprehension to accelerate the reverse engineering process involved in malicious software analysis and classification, as well as for mission assurance. Mr. Charland holds a bachelor and a master’s degree in Computer Science, both from Concordia University, Montreal, Canada.
\endbio

\bio{}
\textbf{Steven H. H. Ding} is an Assistant Professor in the School of Computing at Queen's University, where he leads the L1NNA Artificial Intelligence and Security Lab. His research bridges the domain of machine learning, data mining, and cybersecurity, aiming at addressing critical cybersecurity challenges using AI technologies and securing the future of AI systems. Dr. Ding obtained his Ph.D. from McGill University in 2019, and he was awarded the FRQNT Doctoral Research Scholarship of Quebec and the Dean’s Graduate Award at McGill University.
\endbio

\end{document}